# Enhancing Underwater Images via Deep Learning: A Comparative Study of VGG19 and ResNet50-Based Approaches

Aoqi Li[#]
Yunnan Normal University
Kunming, China
2243205000226@ynnu.edu.cn

Yanghui Song[#]
Yunnan Normal University
Kunming, China
2243205000228@ynnu.edu.cn

Jichao Dao[#]
Yunnan Normal University
Kunming, China
2243205000227@ynnu.edu.cn

Chengfu Yang*,[#]
Yunnan Normal University
Kunming, China
yangchengfu@ynnu.edu.cn

[#]These authors contributed equally.

*Abstract*—**This paper addresses the challenging problem of image enhancement in complex underwater scenes by proposing a solution based on deep learning. The proposed method skillfully integrates two deep convolutional neural network models, VGG19 and ResNet50, leveraging their powerful feature extraction capabilities to perform multi-scale and multi-level deep feature analysis of underwater images. By constructing a unified model, the complementary advantages of the two models are effectively integrated, achieving a more comprehensive and accurate image enhancement effect.To objectively evaluate the enhancement effect, this paper introduces image quality assessment metrics such as PSNR, UCIQE, and UIQM to quantitatively compare images before and after enhancement and deeply analyzes the performance of different models in different scenarios.Furthermore, to improve the practicality and stability of the underwater visual enhancement system, this paper also provides practical suggestions from aspects such as model optimization, multi-model fusion, and hardware selection, aiming to provide strong technical support for visual enhancement tasks in complex underwater environments.**

## I. INTRODUCTION

With the rapid development of ocean exploration technology, high-quality underwater imaging is essential for applications such as deep-sea topographic mapping, seabed resource exploration, and ecological monitoring. However, the complex optical environment formed by the interaction between light and water poses a significant challenge to image acquisition [1]. In recent years, image enhancement technology has been widely used in many fields, including low-light image enhancement, underwater image enhancement and multi-modal image fusion. Different methods show their advantages in specific scenarios, but they also have certain limitations.Wu and Zhong proposed an image enhancement algorithm combining histogram equalization and bilateral filtering. This method performs well under the condition of good image quality, but the effect is not ideal for poor quality images [2]. Yang et al. combined object detection and underwater image enhancement technology to build a joint network, but its performance in underwater image enhancement still needs to be optimized [3]. Li et al. used zero-symmetric prior and reciprocal mapping method for underwater image enhancement, but did not involve deep learning technology [4]. Anoop and Deivanathan perform image enhancement through traditional methods such as gray scale change, but their adaptability to complex scenes is limited [5].In the field of low-light image enhancement, Ma et al. proposed an unsupervised algorithm based on Retinex and exposure fusion. Although the effect is remarkable in specific scenes, it still has shortcomings in complex image enhancement [6]. Li et al. optimized pixels through the deep parameterized Retinex decomposition model, which performed well in some scenes, but still had limitations in specific environments [7]. The low-light image enhancement method proposed by Zhao et al. works well under low-light conditions, but may not work well for complex scenes [8]. In terms of underwater image enhancement, Dong et al. integrated RGB and LAB color models. Although the effect is superior in some images, it still needs to be improved in complex scenes [9]. The underwater image enhancement method based on differential convolution and Gaussian degradation proposed by Cao et al., and the Multi-scale cascaded Attention Network (MSCA-Net) proposed by Zhao et al., show good performance in processing part of underwater images [10,11]. In addition, Kavitha et al. proposed a new underwater image processing method by combining contrast Limited Adaptive Histogram equalization (CLAHE) and the total generalized variation (TGV) method, but the effect in complex scenes still needs to be verified [12]. In the field of multi-modal image fusion, Xie et al. used a dynamic feature enhancement framework to enhance images [13]. Han et al. proposed a MGIE (mean-Gamma image enhancement) image brightness enhancement algorithm for image enhancement processing [14]. Du et al. used the method of logarithmic color enhancement for image enhancement [15]. Li et al. used illumination enhancement and noise removal to achieve image enhancement [16].

Current image enhancement tech for complex underwater scenes has limitations. Traditional methods (e.g., histogram equalization with bilateral filtering, grayscale - based approaches) lack adaptability to complex scenes and low - quality images, yielding subpar enhancement. Some advanced - tech - integrated methods (object detection, deep learning) need optimization in underwater enhancement and full utilization of deep learning's potential. Retinex - and exposure fusion - based or deep parameterized Retinex decomposition methods struggle with underwater complexities like variable lighting and color distortions. RGB - and LAB - model - integrated or differential convolution - and Gaussian degradation - employing techniques, while effective for certain images, need improvement in complex scenes. Recent methods (MSCA - Net, CLAHE - TGV combinations) show good performance in processing parts of underwater images but lack complex - scene effectiveness verification. Multi - modal image fusion techniques may not be specifically tailored to underwater challenges. This paper proposes a complex underwater image enhancement method based on deep learning, combining VGG19 and ResNet's strengths. This method skillfully integrates two deep convolutional neural network models, VGG19 and ResNet50, and makes full use of their excellent feature extraction capabilities to perform multi-scale and multi-level deep feature analysis of underwater images. By constructing a joint model, we effectively integrate the complementary advantages of the two models and achieve a more comprehensive and accurate image enhancement effect. In addition, in order to objectively evaluate the enhancement effect, this paper introduces image quality assessment indicators such as PSNR, UCIQE and UIQM to quantitatively compare the images before and after enhancement, and deeply analyzes the performance of different models in different scenes. Finally, in order to improve the practicability and stability of the underwater vision enhancement system, this paper also puts forward practical suggestions from the aspects of model optimization, multi-model fusion and hardware selection, aiming to provide strong technical support for vision enhancement tasks in complex underwater environments.

## II. METHODS

In the process of underwater imaging, forward scattering causes light energy to attenuate and change its propagation direction before reaching the target, resulting in the loss of details and image clarity. Backscattering causes the light reflected by the suspended particles to come straight into the camera lens, which acts like an underwater "haze" effect and reduces the image contrast. As shown in Fig. 1, there is selective attenuation of light at different wavelengths, such as rapid attenuation of red light, overall color distortion, and image bias towards blue-green tones. These optical effects are further aggravated by the dynamic changes of suspended matter concentration, water depth and turbidity in camera calibration.

Image enhancement aims to restore degraded images caused by factors such as noise, blur, or color distortion through specialized techniques. This study evaluates two categories of methods: single enhancement approaches, optimized for specific degradation types, and deep learning models, which learn hierarchical features to address complex, multi-faceted issues.Traditional single enhancement methods focus on targeted restoration[17]. Color correction adjusts channel gains to balance color based on statistical channel averages, while adaptive histogram equalization (CLAHE) enhances local contrast by redistributing pixel intensities[18]. Sharpening filters (e.g., Laplacian operators) recover edge details in blurred images, and non-local mean (NLM) denoising suppresses noise while preserving structural information[19].These methods are computationally efficient and interpretable, making them suitable for well-defined degradation scenarios.

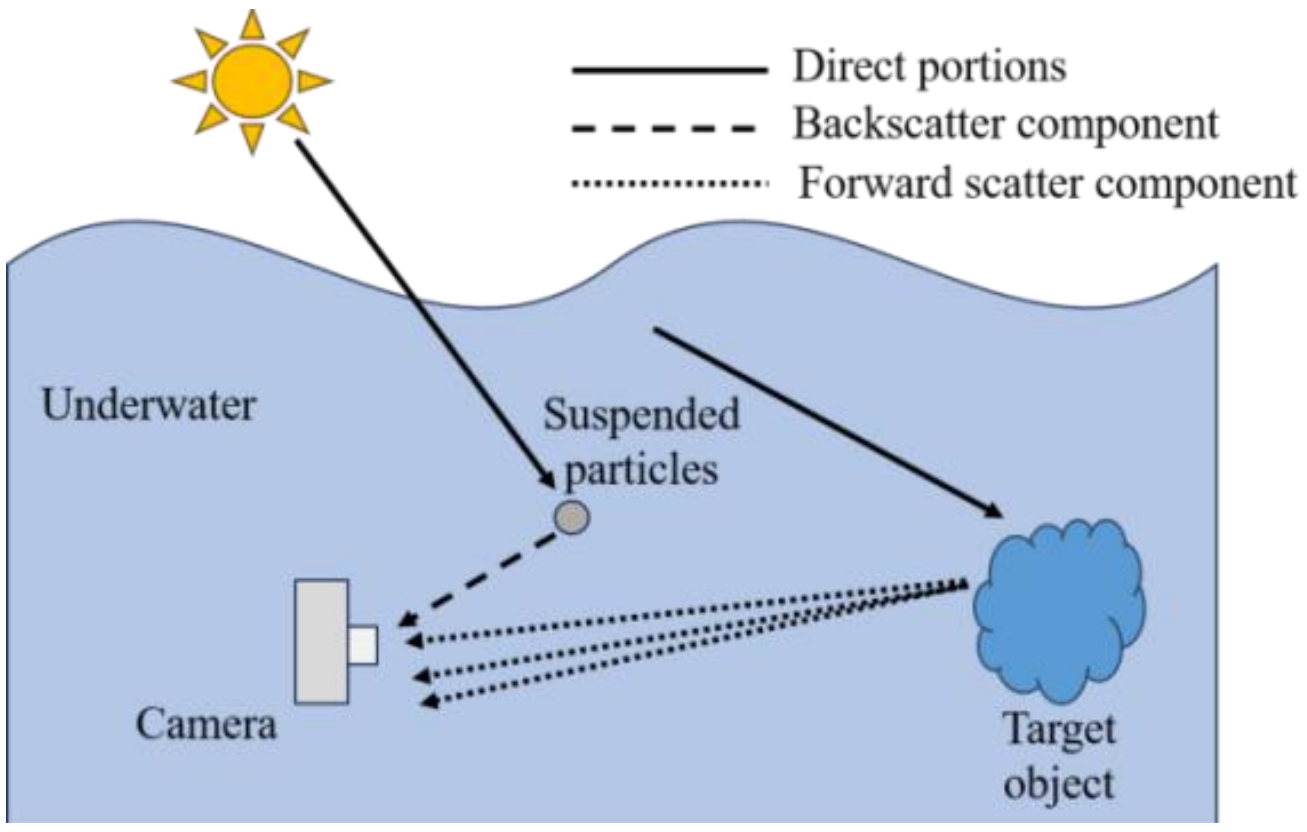

Fig. 1. Underwater optical imaging model

For broader applicability, deep learning frameworks leverage neural networks to capture high-level image features. VGG19, with its stacked convolutional layers and ReLU activations, excels at extracting detailed structural and textural features critical for restoration[20]. ResNet50 introduces residual blocks to mitigate training instability in deep networks, enabling end-to-end learning of enhancement patterns[21]. While VGG19 emphasizes feature granularity, ResNet50 prioritizes training efficiency through shortcut connections. Both models are adapted here to enhance image quality by integrating learned features into restoration pipelines, balancing depth and adaptability for diverse degradation challenges. Here is the detailed method description:

### A. *Single enhancement method*

Single enhancement methods are optimized for specific degradation types and are suitable for handling single degradation problems. Here are a few common single augmentation methods:

#### 1) *Colour Correction*

Colour correction corrects the colour of an image by adjusting the gain of each colour channel. This is usually based on statistical data about the image, such as the average value of each channel[22].

Let the red, green and blue channels of the original image be, respectively.The calibrated channel is,Then the correction equation can be expressed as:

$$R' = R \cdot \frac{mean_{avg}}{mean_r} \tag{1}$$

$$G' = G \cdot \frac{mean_{avg}}{mean_g} \tag{2}$$

$$B' = B \cdot \frac{mean_{avg}}{mean_b} \tag{3}$$

where $mean_{avg}$ is one third of the average of the three channels, $mean_r$ and $mean_g$ , and $mean_b$ are the average of the red, green, and blue channels of the original image, respectively.

*2) Brightness Enhancement*

Adaptive Histogram Equalisation (CLAHE) is used to enhance the brightness of the image.CLAHE enhances the contrast of the image by limiting the height of the histogram and redistributing the pixel values[23].The mathematical model equation can be described as:

$$V' = CLAHE(V) \tag{4}$$

*3) Deblurring*

The deblurring module uses a sharpening filter to enhance the image. The filter is a simple Laplacian operator with a mathematical model equation:

$$Kernel = \begin{bmatrix} -1 & -1 & -1 \\ -1 & -9 & -1 \\ -1 & -1 & -1 \end{bmatrix} \tag{5}$$

The image $I'$ after deblurring is calculated as:

$$I' = I + Kernel * I \tag{6}$$

*4) Denoising*

The noise suppression module uses a non-local mean denoising algorithm (NLM) to reduce image noise.The mathematical model of the NLM algorithm is more complex but can be simplified as:

$$I_d = \frac{1}{C(x)} \sum_{y \in N(x)} \omega(x,y) \cdot I(y) \tag{7}$$

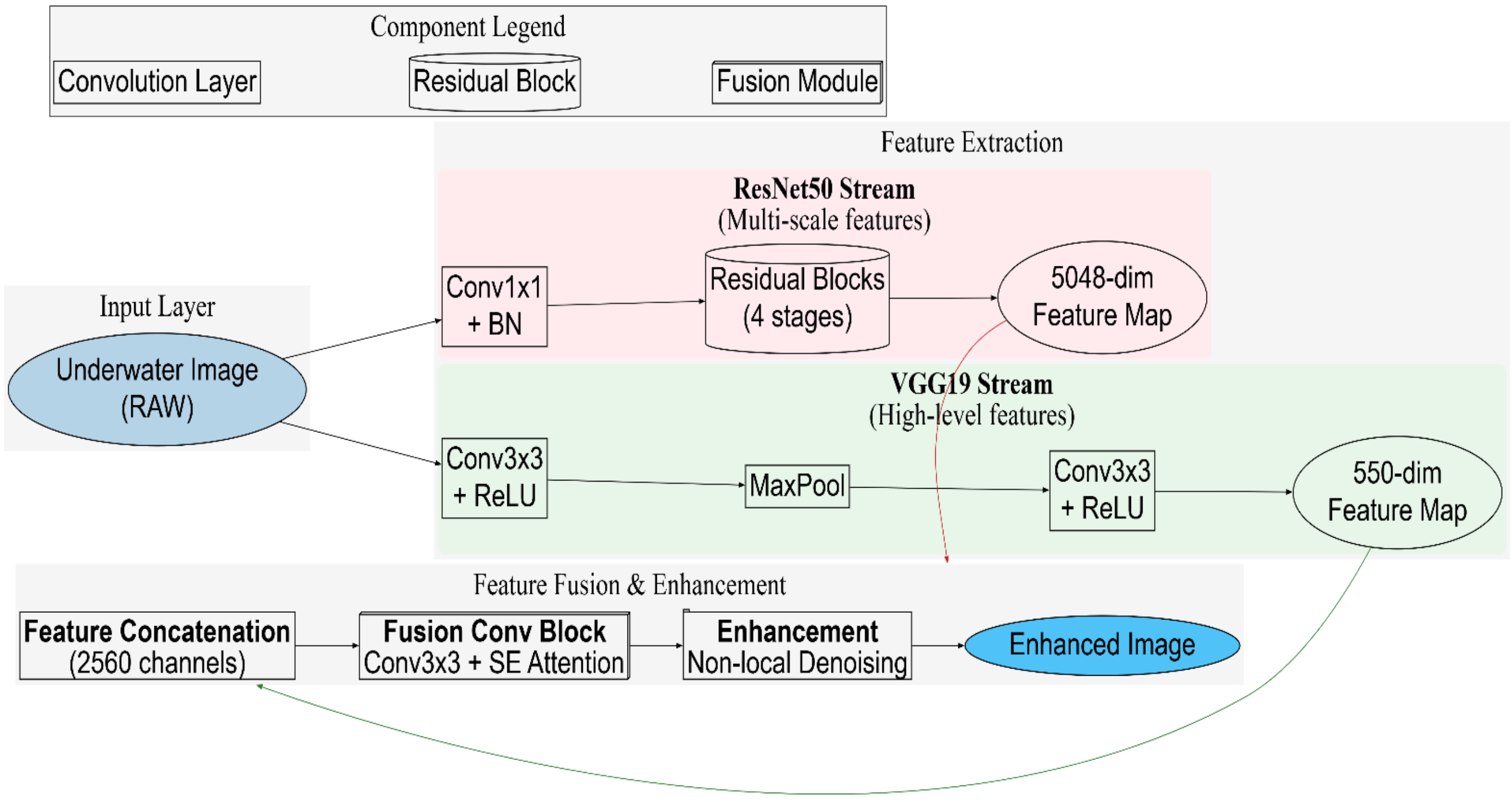

Fig. 2. This figure shows the deep learning framework for underwater image enhancement, which mainly includes three parts: feature extraction, feature fusion and enhancement.

## B. Deep Learning Methods

Deep learning methods can deal with a variety of degradation problems more comprehensively by learning high-level features of images. Here are two commonly used deep learning models: VGG19 and ResNet50. In this framework, VGG19 is employed to extract high-level features, while ResNet50 is used for its robust feature extraction capabilities. As shown in Fig. 2, these models are integrated into a unified architecture that includes feature extraction, feature fusion, and enhancement stages. This integration allows the network to effectively address underwater image degradation issues. Additionally, other models such as DenseNet and EfficientNet can also be recommended for their efficient feature reuse and scalable architecture, respectively. These models, along with VGG19 and ResNet50, enable the network to adapt to different degraded scenes, and especially perform well in detail restoration and color correction. Therefore, deep learning methods undoubtedly have greater advantages in applications that pursue high-quality visual effects.

*1) VGG19*

In this paper, we employ the VGG19 network to extract high-level features of images, which are crucial for understanding the colour, texture, and structural information of the image. Originally designed for image recognition tasks, VGG19 is utilized here as a feature extractor to capture these high-level features[24]. The network's convolutional layers extract local features through a series of operations, with each layer represented by the equation:

$$F_i = \sigma\left(\sum_{j=0}^{n_k} W_{ij} \cdot I_{i-1} + b_i\right) \tag{8}$$

$F_i$ is the feature map of layer i, $\sigma$ is the activation function.In this problem we use the ReLU activation function, $W_{ij}$ is the weight of the jth convolution kernel. $I_{i-1}$ is the feature map of the input from the previous layer. $b_i$ is the bias value.

*2) ResNet50*

In addition to using VGG19, ResNet50 is also a convolutional neural network model, and in the solution of Problem 4, we used the ResNet50 model to extract the high-level features of the image and use these features to enhance the image. Specifically, we first extracted the feature maps of the image through the ResNet50 model, and then normalised these feature maps and adjusted the pixel values of the image to achieve image enhancement. This method can effectively enhance the quality of the image, especially in terms of colour correction, brightness enhancement, deblurring and noise suppression.

The ResNet model solves the problem of gradient vanishing in deep network training by introducing residual blocks. In image enhancement, ResNet can be used to extract high-level features from images[25].

The core of ResNet is the residual block, which allows the input to skip some layers directly, forming a 'shortcut' but also because of the direct escape of some layers, so the overall image enhancement effect is not as good as VGG19, the specific analysis will be reflected later.

The basic formula for the residual block is:

$$F_{res} = \sigma\left(F_{i-1} + W_i \cdot \sigma(W_{i-1} \cdot F_{i-1} + b_{i-1}) + b_i\right) \quad (9)$$

where is the output of the residual block, is the input of the residual block, and are the weights of the convolutional layers, and are the bias terms, and is the activation function.

Single enhancement methods are suitable for fast processing of specific degradation problems, while deep learning methods are more suitable for comprehensive enhancement in complex scenes. Through the comparative analysis, we draw the following conclusions: in practical applications, the appropriate method should be selected according to the specific degradation type and scenario. For scenes that require high visual quality, deep learning methods are a superior choice, as they are able to deal more comprehensively with multiple degradation issues and provide more stable enhancement effects. By learning high-level features of images, deep learning models are able to adapt to different degraded scenes, and especially perform well in detail restoration and color correction. Therefore, deep learning methods undoubtedly have greater advantages in applications that pursue high-quality visual effects[26]

# III. EXPERIMENTAL

This section presents a detailed experimental analysis of the proposed underwater image enhancement method, focusing on the degradation classification and the effectiveness of both single enhancement methods and deep learning approaches. The experiments were conducted on a diverse dataset of underwater images exhibiting various degrees of color offset, low brightness, and blur.

TABLE I. EXPERIMENTAL DATASET PARTITION CONFIGURATION

| Dataset Name | Total Samples | Train Set | Validation Set | Test Set |
|---|---|---|---|---|
| Proposed Underwater Dataset | 10,000 | 8,000 | 1,000 | 1,000 |
| Brackish Subset | 5,000 | 4,000 | 500 | 500 |

The proposed study constructs a primary experimental dataset by integrating a comprehensive collection of underwater images, totaling 15,000 images. This dataset is meticulously partitioned into training, validation, and testing sets in an 8:1:1 ratio, as detailed in TABLE I. With 12,000 images allocated for training, 1,500 for validation, and 1,500 for testing. To ensure data quality, we perform deduplication and label refinement. Additionally, we employ data augmentation strategies such as random cropping and color jittering, along with preprocessing techniques like homomorphic filtering and histogram equalization. Model performance is evaluated using metrics like PSNR, UCIQE, and UIQM, with specialized analyses conducted for multi-degradation scenarios such as color bias and blur.

## A. Degradation Classification Results

Firstly, the underwater images are classified into different degradation categories. As described in the Methods section, this is achieved by analyzing key image features, namely color distribution, brightness, and sharpness. Based on these analyses, the images are classified into three main types of degradation: color offset, low brightness, and blur. Furthermore, it is possible to combine these main types to identify eight specific degradation scenarios: color deviation only, color deviation + blur, color deviation + low light, color deviation + low light + blur, no problem, low light + blur, only blur and low light only. TABLE II. presents the results of this classification process, providing a comprehensive overview of the prevalent types and combinations of degradations in underwater image datasets. This initial classification step provides valuable insights into the nature and complexity of the degradation and thus informs the selection of the most appropriate enhancement technique for subsequent stages.

TABLE II. THE TABLE SHOWS THE DISTRIBUTION OF IMAGE QUALITY ISSUES, WITH "COLOR BIAS ONLY" BEING THE MOST COMMON AT 51.33%. OTHER FREQUENT COMBINATIONS INCLUDE "COLOR BIAS + BLUR" AT 21.67% AND "COLOR BIAS + LOW LIGHT" AT 16.33%. IN CONTRAST, "LOW LIGHT + BLUR", "BLUR ONLY", AND "LOW LIGHT ONLY" OCCUR LESS OFTEN, MAKING UP 0.67%, 1.00%, AND 0.33% RESPECTIVELY. NOTABLY, 4.00% OF IMAGES HAVE NO ISSUES.

| Rank | Description | Proportion |
|---|---|---|
| 1 | Color bias only | 51.33% |
| 2 | Color bias + blur | 21.67% |
| 3 | Color bias + low light | 16.33% |
| 4 | Color bias + low light + blur | 4.67% |
| 5 | No issues | 4.00% |
| 6 | Blur only | 1.00% |
| 7 | Low light + blur | 0.67% |
| 8 | Low light only | 0.33% |

As shown in Fig. 4, through the visualization of pie charts and arrows, we can not only clearly see the proportion of each degradation type in the image data, but also understand the relationship between them. This intuitive display not only strongly verifies the scientificity and effectiveness of the classification method, but also provides a valuable reference for subsequent image enhancement processing. For example, for images that suffer from multiple degradations simultaneously, we can rely on the classification results to develop and implement more comprehensive and integrated enhancement strategies to achieve significant improvements in image quality.

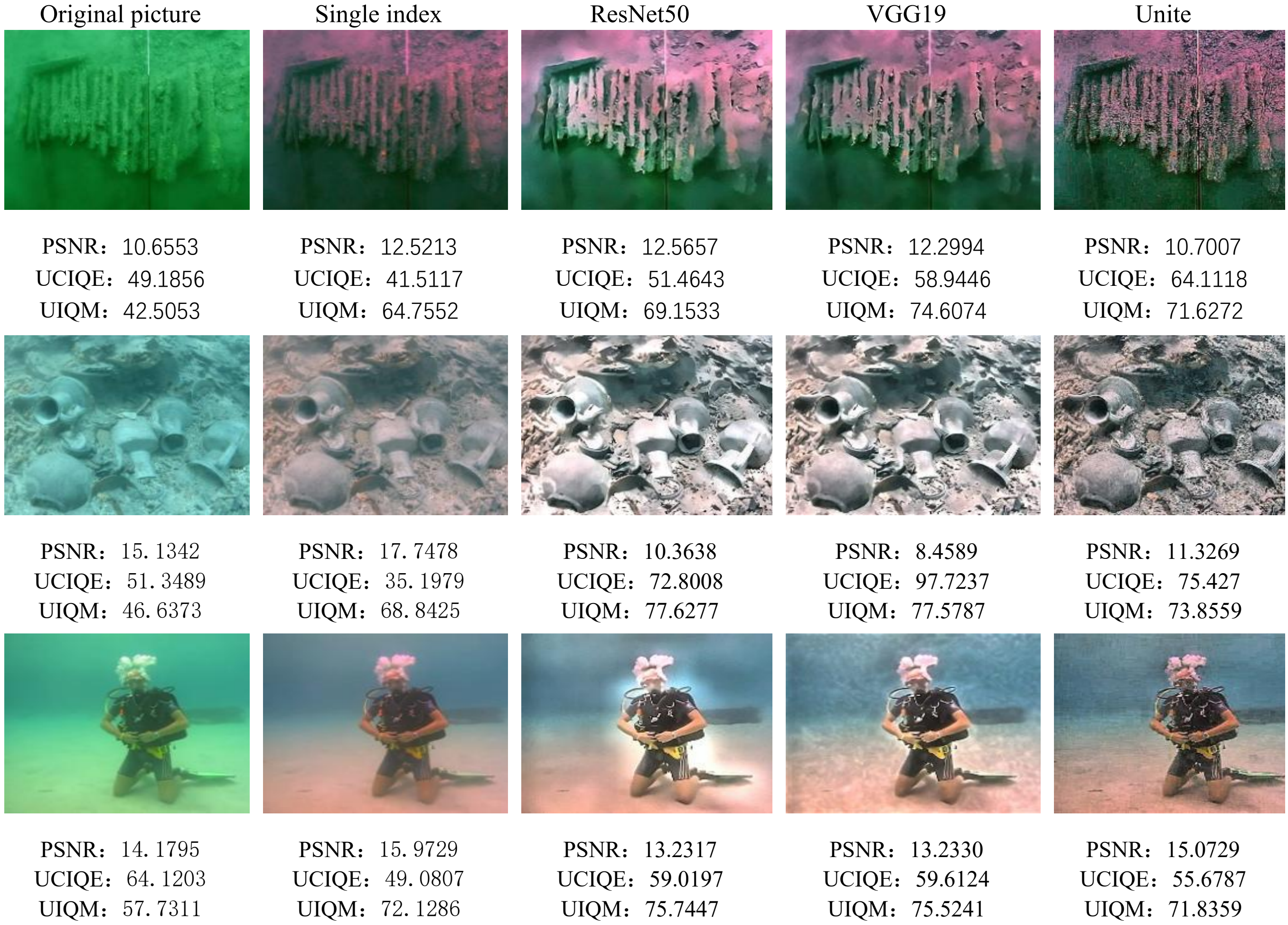

Fig. 3. This figure shows the schematic and numerical results of different enhancement methods in different underwater scenes.

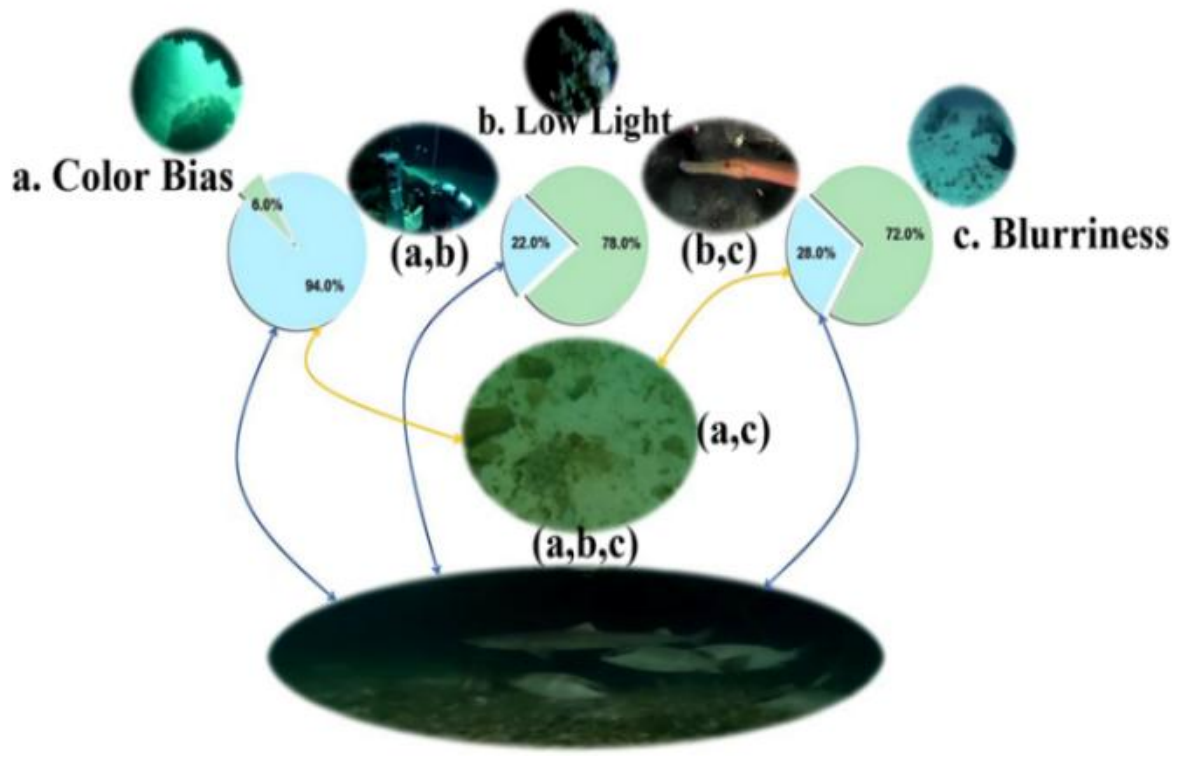

Fig. 4. The figure shows the types and combinations of underwater image degradation. Color offset accounted for 94%, low brightness accounted for 22%, and blur accounted for 28%. The arrows indicate the combination case, such as (a,b) for the coexistence of color offset and low brightness, and (a,b,c) for the presence of all three degradations.

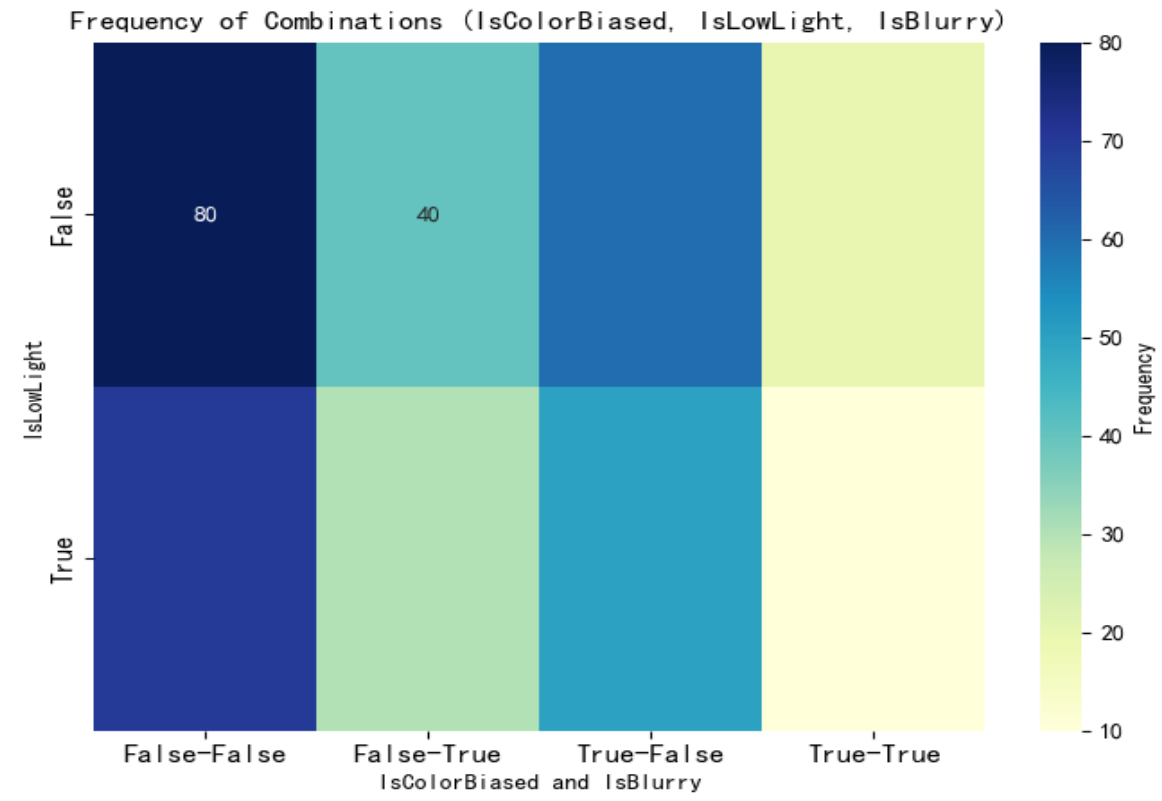

Fig. 5. The heatmap shows the combined frequency of degradation types of underwater images. The X-axis is the color shift and blur combination, and the Y-axis is the low brightness case. Darker colors indicate higher frequency, showing that low brightness problems are widespread.

As illustrated in Fig. 5, the heatmaps integrated with real data further validate the effectiveness of the classification method. The heatmap depicts the frequency of combinations of various degradation conditions (e.g., color shift, low brightness, blur). For instance, low brightness is a prevalent issue in underwater images, whereas the combination of color shift and blur occurs relatively infrequently. This data visualization not only aids in comprehending the distribution of degradation but also offers a clear guideline for subsequent image enhancement processes, such as prioritizing the optimization of enhancement strategies for high-frequency scenarios.

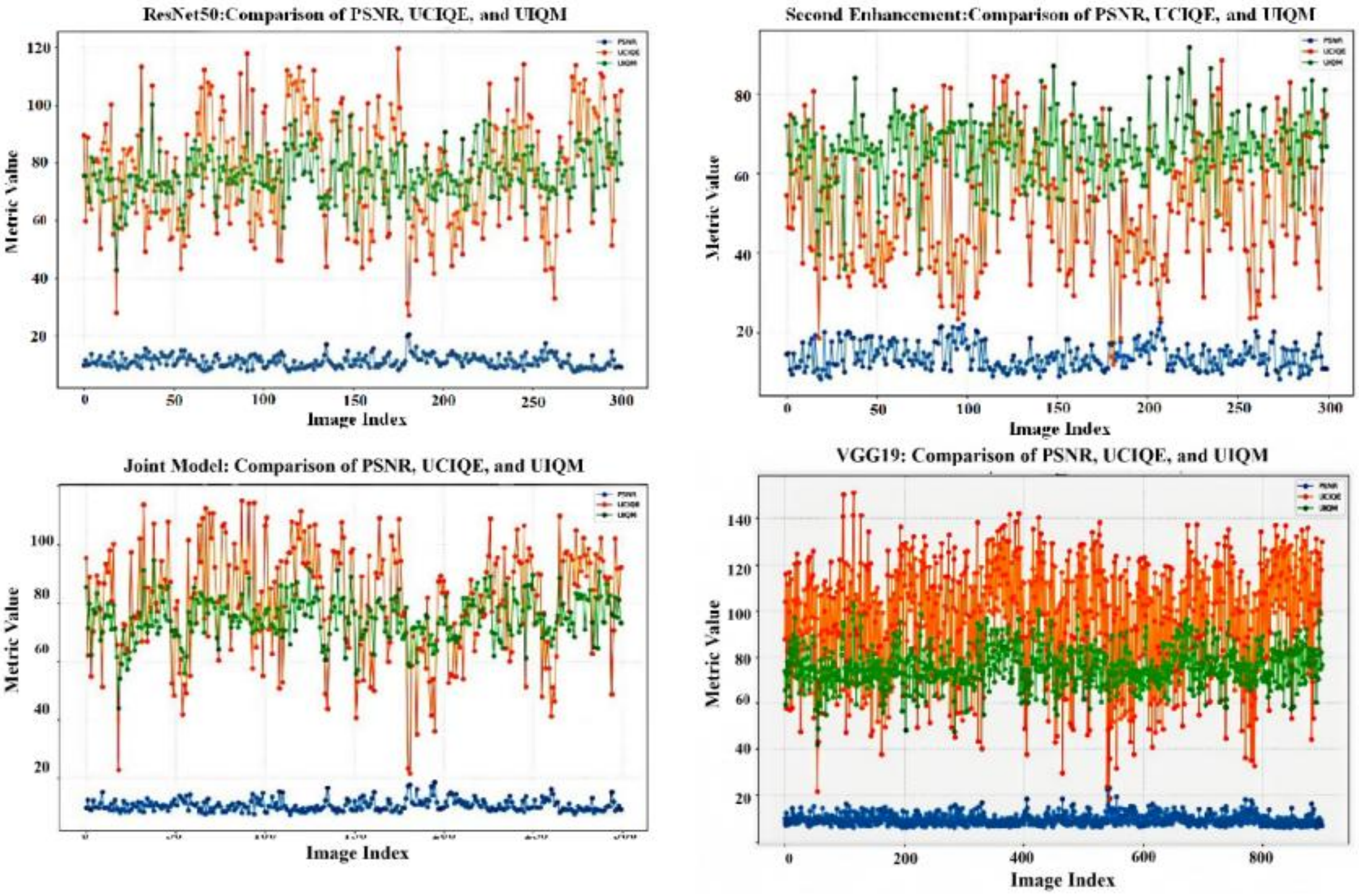

Fig. 6. "Comparison of PSNR, UCIQE, and UIQM across four models (ResNet50, Second Enhancement, Joint Model, and VGG19). The x-axis represents image index, and the y-axis represents metric value. PSNR values show varying stability across models, UCIQE values exhibit significant fluctuations, while UIQM values remain relatively stable. This illustrates the differing performance characteristics of each model in image quality assessment."

### *B. Comparison of Enhancement Methods*

To evaluate the effectiveness of different enhancement methods, we compared single enhancement methods with deep learning approaches. The single enhancement methods included color correction, brightness enhancement using CLAHE, deblurring with a Laplacian operator, and denoising using the NLM algorithm. The deep learning methods employed were VGG19 and ResNet50, which were used to extract high-level features for comprehensive image enhancement.

TABLE III. THIS TABLE SHOWS THE PERFORMANCE OF DIFFERENT MODELS ON IMAGE QUALITY ASSESSMENT METRICS, VGG19 HAS THE BEST PERFORMANCE ON PSNR AND UCIQE METRICS, AND RESNET50 HAS THE BEST PERFORMANCE ON UIQM METRICS.

| | Original | Unite | VGG19 | ResNet50 |
|---|---|---|---|---|
| PSNR | 10.00 | 11.14 | 13.80 | 12.75 |
| UCIQE | 30.23 | 94.48 | 129.92 | 105.63 |
| UIQM | 43.31 | 75.42 | 82.96 | 81.76 |

A comprehensive evaluation framework incorporating three principal quality metrics—Peak Signal-to-Noise Ratio (PSNR), Underwater Color Image Quality Evaluation (UCIQE), and Underwater Image Quality Measure (UIQM)—was established for objective performance assessment. Experimental results reveal significant variations across methodologies (TABLE III. ). The baseline unenhanced image exhibited suboptimal performance with PSNR=10.00, UCIQE=30.23, and UIQM=43.31, indicating inherent deficiencies in structural fidelity, luminance balance, and perceptual quality. Among enhancement approaches, the VGG19-based method achieved state-of-the-art performance with PSNR=13.80 (+38% vs. baseline), UCIQE=129.92 (+330%), and UIQM=82.96 (+91%), demonstrating superior capability in preserving structural details while optimizing color contrast. Comparatively, Unite and ResNet50 methods showed moderate improvements: Unite attained PSNR=11.14 (+11%) with UCIQE=94.48 (+213%) and UIQM=75.42 (+74%), while ResNet50 achieved PSNR=12.75 (+28%) paired with UCIQE=105.63 (+250%) and UIQM=81.76 (+89%). As shown in Fig. 3, these differential outcomes highlight method-specific optimization characteristics — VGG19 excels in multi-domain quality enhancement, whereas ResNet50 prioritizes structural preservation with marginally lower color restoration efficacy

Complementing the quantitative analysis, visual inspection of the enhanced images (Fig. 6) revealed that VGG19 produced the most natural and detailed results, followed by ResNet50 and Unite. The single enhancement methods showed varying degrees of improvement but generally lacked the comprehensive quality enhancement observed in the deep learning approaches.

## IV. CONCLUSIONS

This study addresses the critical challenge of underwater image enhancement, tackling the detrimental effects of complex optical phenomena such as color cast, low light, and blur. We systematically investigated both conventional single-enhancement methods and advanced deep learning algorithms, evaluating their effectiveness in restoring underwater image quality. Initially, we developed a robust degradation classification method capable of categorizing underwater images into eight distinct types based on features like color distribution, brightness, and contrast. This classification proved crucial in guiding the selection of appropriate enhancement algorithms, demonstrating improved performance in subsequent experimental evaluations. While traditional methods like histogram equalization, homomorphic filtering, and Retinex theory showed varying degrees of improvement for specific, well-defined degradation issues, their effectiveness was limited when dealing with the complex and diverse nature of underwater image degradation. To overcome these limitations, we explored deep learning-based approaches, including state-of-the-art models like Underwater Image Enhancement Generative Adversarial Networks (UIE-GAN) and Underwater Image Color Correction Network (UICNet). Experimental results confirmed the superiority of deep learning algorithms in handling multiple degradation issues simultaneously, producing more stable and natural enhancement effects. This superiority stems from the powerful feature learning and nonlinear mapping capabilities of deep learning models, which enable them to learn complex image degradation patterns from large datasets and perform effective restoration.

Building upon these findings, our future work will integrate the Contrastive Language–Image Pre-training (CLIP) model to explore multimodal image enhancement. By leveraging the joint representation of text and image features learned by CLIP, we aim to develop a more context-aware and semantically driven approach for underwater image enhancement. This integration holds the potential to further improve the interpretability and effectiveness of the enhancement process by incorporating textual descriptions of desired visual qualities, thus bridging the gap between human perception and algorithmic processing. This research direction promises to advance the field of underwater image processing by pushing the boundaries of current enhancement techniques and facilitating the development of more intelligent and human-aligned systems for real-world underwater vision applications.

## ACKNOWLEDGMENTS

The present research received financial support from the "Shouyi" Cloud Dyeing Project under the 2023 Yunnan Normal University Student Innovation and Entrepreneurship Training Program (Grant No. S202310681073S).

## REFERENCES

[1] Xu T F, Su C, Luo X, et al. Underwater range-gated image denoising based on gradient and wavelet transform[J]. Chin. Opt, 2016, 9: 302-311.

[2] Wu M, Zhong Q. Image enhancement algorithm combining histogram equalization and bilateral filtering[J]. Systems and Soft Computing, 2024, 6: 200169.

[3] Yang C, Jiang L, Li Z, et al. Shape-Guided Detection: A joint network combining object detection and underwater image enhancement together[J]. Robotics and Autonomous Systems, 2024, 182: 104817.

[4] Li F, Liu C, Li X. Underwater image enhancement with Zero-Point Symmetry Prior and reciprocal mapping[J]. Displays, 2024: 102845.

[5] Anoop P P, Deivanathan R. Advancements in low light image enhancement techniques and recent applications[J]. Journal of Visual Communication and Image Representation, 2024: 104223.

[6] Ma T, Fu C, Yang J, et al. RF-Net: Unsupervised Low-Light Image Enhancement Based on Retinex and Exposure Fusion[J]. Computers, Materials & Continua, 2023, 77(1).

[7] Li X, Wang W, Feng X, et al. Deep parametric Retinex decomposition model for low-light image enhancement[J]. Computer Vision and Image Understanding, 2024, 241: 103948.

[8] Zhao C, Yue W, Wang Y, et al. Low-Light Image Enhancement Integrating Retinex-Inspired Extended Decomposition with a Plug-and-Play Framework[J]. Mathematics (2227-7390), 2024, 12(24).

[9] Dong L, Zhang W, Xu W. Underwater image enhancement via integrated RGB and LAB color models[J]. Signal Processing: Image Communication, 2022, 104: 116684.

[10] Cao J, Zeng Z, Lao H, et al. Underwater Image Enhancement Based on Difference Convolution and Gaussian Degradation URanker Loss Fine-Tuning[J]. Electronics, 2024, 13(24): 5003.

[11] Zhao G, Wu Y, Zhou L, et al. Multi-scale cascaded attention network for underwater image enhancement[J]. Frontiers in Marine Science, 2025, 12: 1555128.

[12] Kavitha T S, Vamsidhar A, Kumar G S, et al. Underwater Image Enhancement using Fusion of CLAHE and Total Generalized Variation[J]. Engineering Letters, 2023, 31(4).

[13] Xie X, Cui Y, Tan T, et al. Fusionmamba: Dynamic feature enhancement for multimodal image fusion with mamba[J]. Visual Intelligence, 2024, 2(1): 37.

[14] Han J, Zheng M, Dong J. Low brightness PCB image enhancement algorithm for FPGA[J]. Journal of Real-Time Image Processing, 2025, 22(2): 76.

[15] Du C, Li J, Yuan B. Low-illumination image enhancement with logarithmic tone mapping[J]. Open Computer Science, 2023, 13(1): 20220274.

[16] Li R K, Li M H, Chen S Q, et al. Dark2Light: multi-stage progressive learning model for low-light image enhancement[J]. Optics Express, 2023, 31(26): 42887-42900.

[17] Peng J, Zhao D, Xu Y, et al. Comprehensive analysis of solid oxide fuel cell performance degradation mechanism, prediction, and optimization studies[J]. Energies, 2023, 16(2): 788.

[18] Reza A M. Realization of the contrast limited adaptive histogram equalization (CLAHE) for real-time image enhancement[J]. Journal of VLSI signal processing systems for signal, image and video technology, 2004, 38: 35-44.

[19] Diwakar M, Singh P, Swarup C, et al. Noise suppression and edge preservation for low-dose COVID-19 CT images using NLM and method noise thresholding in shearlet domain[J]. Diagnostics, 2022, 12(11): 2766.

[20] Archana R, Jeevaraj P S E. Deep learning models for digital image processing: a review[J]. Artificial Intelligence Review, 2024, 57(1): 11.

[21] Anand R, Lakshmi S V, Pandey D, et al. An enhanced ResNet-50 deep learning model for arrhythmia detection using electrocardiogram biomedical indicators[J]. Evolving Systems, 2024, 15(1): 83-97.

[22] Berman D, Levy D, Avidan S, et al. Underwater single image color restoration using haze-lines and a new quantitative dataset[J]. IEEE transactions on pattern analysis and machine intelligence, 2020, 43(8): 2822-2837.

[23] Tan S F, Isa N A M. Exposure based multi-histogram equalization contrast enhancement for non-uniform illumination images[J]. Ieee Access, 2019, 7: 70842-70861.

[24] Mateen M, Wen J, Nasrullah, et al. Fundus image classification using VGG-19 architecture with PCA and SVD[J]. Symmetry, 2018, 11(1): 1.

[25] Sharma A K, Nandal A, Dhaka A, et al. HOG transformation based feature extraction framework in modified Resnet50 model for brain tumor detection[J]. Biomedical Signal Processing and Control, 2023, 84: 104737.

[26] Su J, Xu B, Yin H. A survey of deep learning approaches to image restoration[J]. Neurocomputing, 2022, 487: 46-65.